\documentclass[letterpaper]{article} 
\usepackage{aaai25}  
\usepackage{times}  
\usepackage{helvet}  
\usepackage{courier}  
\usepackage[hyphens]{url}  
\usepackage{graphicx} 
\usepackage{natbib}  
\usepackage{caption} 
\usepackage{algorithm}

\usepackage{newfloat}
\usepackage{listings}
\DeclareCaptionStyle{ruled}{labelfont=normalfont,labelsep=colon,strut=off} 
\floatstyle{ruled}
\newfloat{listing}{tb}{lst}{}
\floatname{listing}{Listing}
\nocopyright 

\usepackage{makeidx}  
\usepackage{graphicx}
\usepackage{xcolor}
\usepackage[noend]{algpseudocode}
\usepackage{subcaption}
\usepackage{amssymb}
\usepackage{amsmath}
\usepackage{wasysym}
\usepackage{hhline}

\usepackage{xspace}
\newcommand{\ourMethod}[1][]{\text{Comp-LTL#1}\xspace}

\newcommand{\booleanComposition}[1][]{\text{BC#1}\xspace}
\newcommand{\rewardMachines}[1][]{\text{RM#1}\xspace}

\newcommand \LTLalways{\square}

\newcommand \LTLeventually{\lozenge} 
\newcommand \LTLand{\wedge}
\newcommand \LTLor{\vee}
\newcommand \LTLuntil{\mathcal{U}}
\newcommand \Not{\mathopen{\neg}}

\newtheorem{sketch}{Proof (sketch)}
\newtheorem{definition}{Definition}
\newtheorem{problem}{Problem}
\newtheorem{remark}{Remark}
\newtheorem{theorem}{Theorem}

\usepackage{tikz}
\usetikzlibrary{automata, positioning, arrows}

\tikzset{
->, 
>=stealth', 
node distance=2cm, 
every state/.style={thick}, 
initial text=$ $, 
}

\title{\ourMethod: Temporal Logic Planning via Zero-Shot Policy Composition}
\author{
    Taylor Bergeron\textsuperscript{\rm 1}, Zachary Serlin\textsuperscript{\rm 2}, and Kevin Leahy\textsuperscript{\rm 1}
}
\affiliations{
    \textsuperscript{\rm 1}Worcester Polytechnic Institute\\
    100 Institute Rd\\
    Worcester, MA 01609 USA\\
    \textsuperscript{\rm 2}MIT Lincoln Laboratory\\
    244 Wood St\\
    Lexington, MA 02421 USA\\
}

\begin{document}

\maketitle

\begin{abstract}
This work develops a zero-shot mechanism, \ourMethod, for an agent to satisfy a Linear Temporal Logic (LTL) specification given existing task primitives trained via reinforcement learning (RL). Autonomous robots often need to satisfy spatial and temporal goals that are unknown until run time. Prior work focuses on learning policies for executing a task specified using LTL, but they incorporate the specification into the learning process. Any change to the specification requires retraining the policy, either via fine-tuning or from scratch. We present a more flexible approach--to learn a set of composable task primitive policies that can be used to satisfy arbitrary LTL specifications without retraining or fine-tuning. Task primitives can be learned offline using RL and combined using Boolean composition at deployment. This work focuses on creating and pruning a transition system (TS) representation of the environment in order to solve for deterministic, non-ambiguous, and feasible solutions to LTL specifications given an environment and a set of task primitive policies. We show that our pruned TS is deterministic, contains no unrealizable transitions, and is sound. We verify our approach via simulation and compare it to other state of the art approaches, showing that \ourMethod is safer and more adaptable.
\end{abstract}

%

\section{Introduction}

A major goal in autonomous systems is the deployment of robots that are capable of executing complex tasks in safety-critical scenarios, such as search and rescue (SAR), transportation, and healthcare emergency services~\cite{bogue2019,bravo2015}. In these situations, the requirements may be time-varying, interdependent, and otherwise complex. 
For example, in healthcare, it is imperative to have a flexible ``triage" approach to high priority tasks while still safely executing actions. Additionally, in a fire SAR scenario, certain rooms may become inaccessible, requiring a change of the environment representation and temporal task ordering for searching the building. 

One approach to addressing such complex task executions is planning with linear temporal logic (LTL)~\cite{baier2008,kressgazit2008}. LTL allows a user to specify tasks with complex temporal and inter-task relationships. 
A major strength of this approach is the focus on correct-by-construction algorithms that are capable of planning for an arbitrary formula specified by a user. However, many associated planning approaches require reliable task models 
in order to guarantee satisfaction of an LTL specification \cite{kressgazit2008,beltabook2017}.

Recent work has focused on using reinforcement learning (RL) to meet LTL specifications when a model of the system is unavailable.
Many of these approaches incorporate an automaton or similar structure in the training process.
One such approach is Reward Machines~\cite{icarte2018} (\rewardMachines), which encodes the specification as reward functions while exposing the reward function structure to the training agent. Other approaches use an automaton to guide the learning process in the presence of partially-satisfiable LTL specifications~\cite{cai2023}, or to learn sub-policies corresponding to edges in an automaton~\cite{li2019}.
While all of these approaches learn policies that are capable of executing a high-level task specified using LTL, they incorporate the specification into the learning process, so any change to the specification requires retraining the policy.
We desire a more flexible approach -- to learn a set of policies that can be used to satisfy arbitrary specifications without retraining.

A closely related approach is Skill Machines ~\cite{tasse2024skill}, which leverages prior work on zero-shot composition~\cite{tasse2020booleantaskalgebrareinforcement}, to satisfy a proposition on a reward machine. While changing the specification does not require re-training from scratch, it nonetheless requires fine-tuning of the policies to guarantee satisfaction.
In our work, we aim to find a solution that requires no retraining and no fine-tuning beyond the initial training of tasks, while still being able to satisfy an arbitrary LTL specification. 
Another closely related approach, LTL-Transfer ~\cite{liu2024ltltransferskilltransfertemporal}, is a zero-shot LTL solution that adheres to safety specifications encoded in the LTL formula. LTL-Transfer creates a B\"uchi automaton representation of the specification, and trains on the transitions in the automaton and transfers the logic to new LTL specifications. This constrains the possibilities of LTL specifications that can be satisfied to the set of automaton transitions that have already been explored during the training pass, whereas we desire a more broadly applicable solution.

\par
\begin{figure*}[h]
\centering
\includegraphics[width=1.0\textwidth]{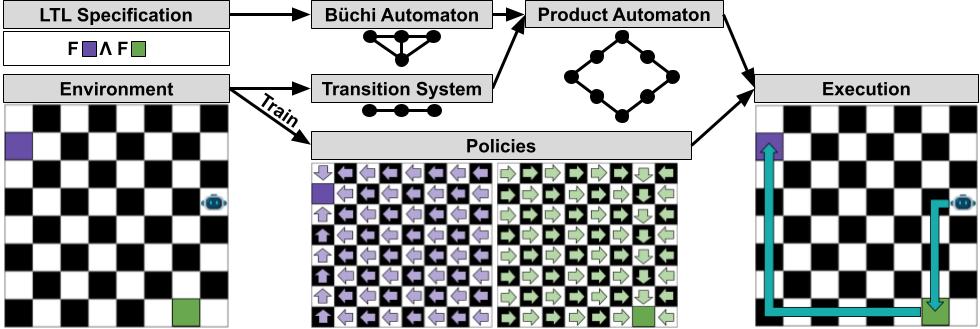}
\caption{\ourMethod training and path execution pipeline. Policies are trained for a set of base tasks. The environment is abstracted as a policy-aware transition system, which can then be composed with a B\"{u}chi automaton to plan over the Boolean composition of task policies.}
\label{ourTrainingPipeline}
\end{figure*}

In this work, we propose a framework for finding a satisfying solution for an environment and specification regardless of the exact environment, specification, or policies.
Inspired by~\citet{kloetzer2008fully} and recent work in zero-shot Boolean Composition~\cite{tasse2020booleantaskalgebrareinforcement} (\booleanComposition), we observe that compositional approaches allow us to satisfy Boolean constraints on automaton representations of LTL specifications. We leverage our prior work on safety-aware Boolean compositions of primitive policies to ensure the solution can be run zero-shot~\cite{leahy2023safetyaware}, and that the satisfying word can be achieved in the environment.
Figure~\ref{ourTrainingPipeline} shows our approach, \ourMethod.

The specific contributions of this work are the following:

\begin{enumerate}
    \item We develop a method for abstracting a geometric representation of an environment into a transition system (TS) with transition labels representing feasible Boolean combinations of tasks to transition between regions;
    \item We resolve nondeterminism in the transitions enabled by the Boolean composition of base task policies; and
    \item We demonstrate that this representation allows zero-shot satisfaction of LTL specifications at run time.
\end{enumerate}

We support our theoretical results with case studies in simulation and comparison to other approaches.

\section{Background and Problem Formulation}
\label{sec:background}

\par
We consider an agent moving in a planar environment according to a high-level mission description. 
We denote the agent's environment $E\subseteq\mathbb{R}^2$. The environment contains non-intersecting regions $R\subseteq E$ taking labels from $2^\Sigma$, where $\Sigma$ is a set of atomic propositions corresponding to properties of interest in the environment (see Fig~\ref{fig:env}). We define a labeling function $L:\mathcal{R}\rightarrow 2^\Sigma$, that defines which regions are labeled with which properties, where $\mathcal{R}$ is the set of all labeled regions. For $\sigma\in\Sigma$, we assume an agent can perform task $\sigma$ if it is in a region $r$ of the environment such that $\sigma\in L(r)$. Given an agent's motion through a sequence of regions $r_0,r_1,\ldots$, the agent produces a word $\tau = L(r_0),L(r_1)\ldots$.

The agent's task is specified using linear temporal logic (LTL)~\cite{baier2008}. 
LTL includes Boolean operators, such as \texttt{AND} ($\LTLand$), \texttt{OR} ($\LTLor$), and \texttt{NOT} ($\Not$), along with time based operators \emph{eventually} ($\LTLeventually$), \emph{always} ($\LTLalways$), and \emph{until} ($\LTLuntil$). 
The formal syntax of LTL in Backus–Naur form is
\begin{equation}
    \phi ::= \top | \sigma | \Not \sigma | \phi_1 \LTLor \phi_2 | \phi_1 \LTLand \phi_2 | \phi_1 \LTLuntil \phi_2 | \LTLeventually \phi_1 | \LTLalways \phi_1\:,
\end{equation}
where $\sigma \in \Sigma$ is an atomic proposition, and $\phi$, $\phi_1$, and $\phi_2$ are LTL formulas~\cite{baier2008}. 

Due to space constraints, we do not describe the semantics of LTL here, but provide a brief intuition. A sequence $\tau$ satisfies a specification $\phi$ (written $\tau\models\phi$) if the sequence matches the properties specified by $\phi$. For example, if $\phi=\lozenge \sigma$ (``eventually $\sigma$"), $\tau\models\phi$ if $\sigma$ occurs at some point in $\tau$. Similarly, $\square\sigma$ (``always $\sigma$") requires that $\sigma$ appear at every point in $\tau$.
Interested readers are directed to~\citet{baier2008} for more details on the semantics of LTL.
Importantly, off-the-shelf software, such as SPOT, can automatically translate LTL specifications into B\"{u}chi automata~\cite{spot}. Furthermore, each transition on such automata can be described by 
a Boolean combination of atomic propositions.

\par
We assume there is no transition model for the agent in its environment, and therefore we use
reinforcement learning (RL) to learn a model for the agent to execute its tasks. RL is a branch of machine learning that maps states to actions in order to maximize a reward function~\cite{suttonbarto1998}. 
To facilitate satisfaction of temporal logic objectives, we will leverage our prior work~\cite{leahy2023safetyaware} on safety-aware task composition to train policies for a given set of tasks. In that work, we proposed a method for learning policies that have ``minimum-violation" (MV) safety semantics. 
For a word $\tau$, let the number of positions in the word with non-empty symbols be denoted $\vert \tau\vert$ and the set of symbols in the last position of the word be denoted $\tau_{f}$. Then, for a Boolean formula $\varphi$, we define minimum-violation semantics as follows.
\begin{definition}
Minimum-violation Path: A word $\tau$ is a minimum-violation path if $\vert\tau\vert > 1$ and $\tau_{f} \models \varphi$ and there is no word $\tau'$ such that $\vert\tau' \vert < \vert\tau\vert$~\cite{leahy2023safetyaware}.
\end{definition}
Intuitively, a minimum-violation is a path that: 1) terminates in a state that satisfied a Boolean formula; 2) if possible, visits no additional labeled states; and 3) if not possible, visits the fewest additional labeled states.

\begin{problem}
Given a set of labels $\Sigma$, an environment $E$ labeled from $\Sigma$, train policies for achieving tasks corresponding to $\sigma\in\Sigma$ such that the policies can be used to satisfy an LTL formula $\phi$ over $\Sigma$ without additional training.
\label{prob1}
\end{problem}

\section{Technical Approach}
\label{sec:approach}


\begin{figure*}[h]
\includegraphics[width=1.0\textwidth]{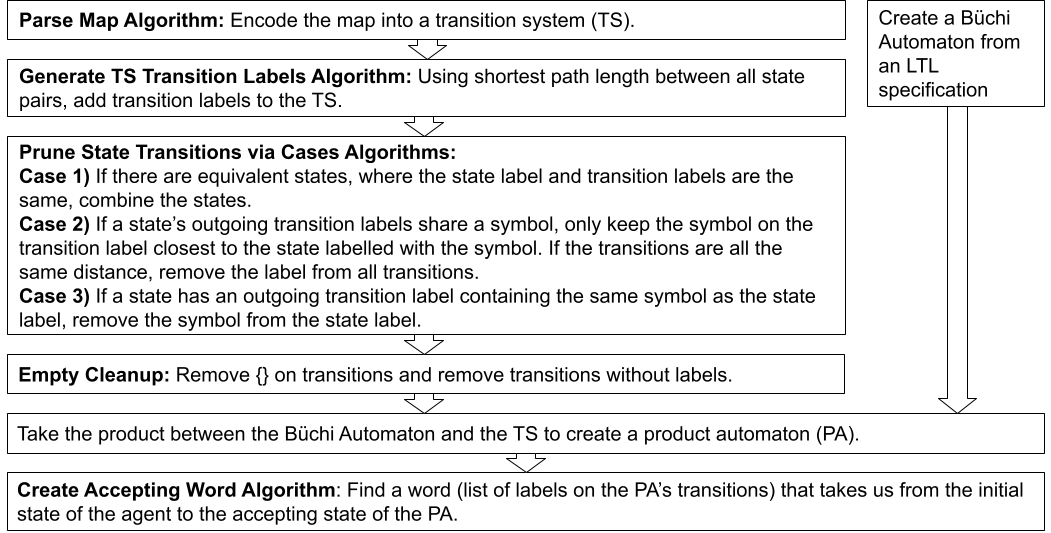}
\caption{Block diagram representation of the overall Zero-Shot Algorithm for \ourMethod. It includes the TS generation algorithm described in Section ~\ref{sec:generate_ts}, from parsing the map, generating the initial TS, and pruning the TS to remove equivalency, solve ambiguity, and ensure feasibility. All sub algorithms are denoted in bold.}
\label{overviewBlockDiagram}
\end{figure*}

\par
To solve Problem~\ref{prob1} we introduce a novel 
policy-aware environment abstraction as described below.
Figure~\ref{overviewBlockDiagram} shows an overview of the proposed solution algorithms. 
First, we create a TS~\cite{beltabook2017} that captures both the topology of the environment as well as the policies for each base task to move the agent between regions. 
Constructing such a TS is conservative and can introduce ambiguity and non-determinism. To that end, we identify 3 cases for pruning the edge labels to remove non-determinism due to the reliance on task composition. 
The resulting TS can be used for planning to satisfy an LTL specification in the standard method~\cite{beltabook2017}, while accurately capturing the behavior created by the RL policies.

\subsection{Generating The Transition System}
\label{sec:generate_ts}

To facilitate reasoning about satisfaction of an LTL specification, we abstract the environment as a transition system (TS). A
TS describes the discrete behavior of a system via states and transitions 
and is formally defined as follows.
\begin{definition}
    A \emph{transition system} (TS) is a tuple, $TS = (S, Act, \rightarrow, I, AP, L)$, where
    \begin{itemize}
        \item $S$ is a finite set of states;
        \item $Act$ is a finite set of actions;
        \item $\rightarrow\subseteq S\times Act\times S$ is a transition relation;
        \item $I\in S$ is an initial state;
        \item $AP$ is a set of atomic propositions; and
        \item $L:S\rightarrow 2^{AP}$ is a labeling function.
    \end{itemize}
\end{definition}
We do not assume that the TS is deterministic. A \emph{deterministic} transition system is a TS in which the transition relation 
$\rightarrow \subseteq S \times Act \times S$ is deterministic. 
For every $\left(s_i,\alpha_i,s_{i+1}\right)\in \rightarrow$,
given $s_i$ and $\alpha_{i+1}$, $s_{i+1}$ is unique.

\begin{figure}[h]
\centering
\begin{subfigure}[b]{.35\linewidth}
    \centering
\resizebox{\linewidth}{!}{
\begin{tikzpicture}
    \draw[draw=black] (-1,-1) rectangle ++  (12,12);
    \draw plot[smooth cycle] coordinates {(1, 2) (2, 2) (2, 1) (1,1)};
    \node at (1.5,1.5) {\huge $a$};
    \draw plot[smooth cycle] coordinates {(7.5,9.5) (9.5, 9.5) (9.5, 7.5) (7.5, 7.5)};
    \node at (8.5,7.5) {\huge $a$};
    \draw plot[smooth cycle] coordinates {(8,9) (9, 9) (9, 8) (8, 8)};
    \node at (8.5,8.5) {\huge $c$};

    \draw plot[smooth cycle] coordinates {(5.5,4.5) (9.5, 4.5) (9.5, .5) (5.5, .5)};
    \node at (7.5,.5) {\huge $a$};
    \draw plot[smooth cycle] coordinates {(6.5,3.5) (8.5, 3.5) (8.5, 1.5) (6.5, 1.5)};
    \node at (7.5,2.5) {\huge $c$};
    
    \draw plot[smooth cycle] coordinates {(4.5,5.5) (4.5, 9.5) (.5, 9.5) (.5, 5.5)};
    \node at (2.5,5.5) {\huge $a$};
    \draw plot[smooth cycle] coordinates {(3.5,6.5) (3.5, 8.5) (1.5, 8.5) (1.5, 6.5)};
    \node at (2.5,7.5) {\huge $b$};   
\end{tikzpicture}
}\caption{}\label{fig:env}
\end{subfigure}
\hfill
\begin{subfigure}[b]{.6\linewidth}
    \centering
    \resizebox{\linewidth}{!}{
\begin{tikzpicture}
\node[state,label={above:$\lbrace b\rbrace$}] (q0) {$q_0$};
        \node[state,right of=q0,label={above:$\lbrace a\rbrace$}] (q1) {$q_1$};
        \node[state,right of=q1,label={above:$\lbrace\rbrace$}] (q2) {$q_2$};
        \node[state,right of=q2,label={above:$\lbrace a\rbrace$}] (q3) {$q_3$};
        \node[state,right of=q3,label={above:$\lbrace c\rbrace$}] (q4) {$q_4$};

        \node[state,above of=q2,label={above:$\lbrace a\rbrace$}] (q5) {$q_5$};
        
        \node[state,below of=q2,label={above:$\lbrace a\rbrace$}] (q6) {$q_6$};
        \node[state,left of=q6,label={above:$\lbrace c\rbrace$}] (q7) {$q_7$};
        
        \draw 
        (q0) edge[bend right, below] (q1)
        (q1) edge[bend right, below] (q2)
        (q2) edge[bend right, below] (q3)
        (q3) edge[bend right, below] (q4)
        
        (q1) edge[bend right, above] (q0)
        (q2) edge[bend right, above] (q1)
        (q3) edge[bend right, above] (q2)
        (q4) edge[bend right, above] (q3)
        
        (q6) edge[bend right, below right] (q2)
        (q7) edge[bend right, right] (q6)
        (q2) edge[bend right, below left](q6)
        (q6) edge[bend right, left] (q7)

        (q5) edge[bend right, above left](q2)
        (q2) edge[bend right, above right](q5);
\end{tikzpicture}}\caption{}\label{fig:init_ts}
\end{subfigure}
\caption{Example of an environment with regions to be parsed by Algorithm~\ref{alg:generate_ts}~(\ref{fig:env}). A corresponding TS~(\ref{fig:init_ts}) cannot be used directly, since it is not known a priori which region with a given label will be reached when executing a policy trained with RL.}
\label{example8}
\end{figure}
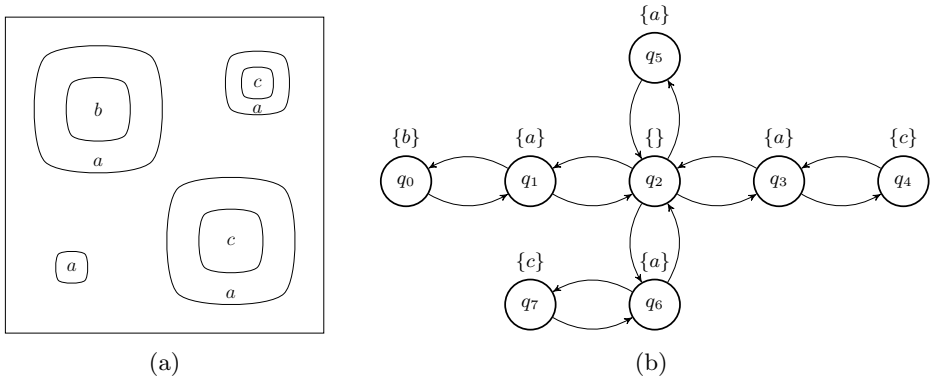

To create the initial TS, each region is instantiated as a state, and adjacent regions are connected by transitions. This captures the topology of the environment. In planning- and control-based approaches, it is typical to assume that an agent can travel between any adjacent regions. For example, Fig.~\ref{fig:env} shows an environment and a corresponding TS~(\ref{fig:init_ts}). In a planning framework, an agent may choose which of the regions labeled $a$ to visit. Using our RL approach, however, for an agent in the unlabeled region $q_2$, executing a policy corresponding to $a$ may cause the agent to visit $q_1$, $q_3$, $q_5$, or $q_6$, since the transition function is unknown. We introduce a pruning process to model and resolve such ambiguities.

First, our TS must accurately reflect the results of applying a given task policy from each state. Algorithm~\ref{alg:generate_ts} generates the transition labels for the TS for this purpose. In this and the following algorithms, $S$ is the set of states, $s \subseteq S$ is a state, $T$ is the set of transitions, and $t \subseteq T$ is a transition. For each transition, the algorithm checks the distance (represented as function $d(\cdot)$) between the start and end state of that transition to each other state in the TS (lines 3--6). If the distance from the start state to a state is greater than the distance from the end state to a state, the transition is approaching that state (and its associated label), so that state's label is added to the transition (lines 7--8).

\begin{algorithm}[h]
  \caption{Generate Transition System Transition Labels}
  \begin{algorithmic}[1]
\Procedure{GenerateTSLabels}{$S, T$}
\State initialize transition labels $L$
\For{$t \in T$} \Comment{t=(startState, endState)}
    \For{$s \in S$}
        \State$d_{start} = d(startState,s)$ 
        \State$d_{end} = d(endState,s)$
        \If{\texttt{$d_{start} > d_{end}$}}
            \State$L[t] \gets L[t] \cup L[s]$
        \EndIf
    \EndFor
\EndFor
\State \Return $L$\Comment{The transition system labels}
\EndProcedure
\end{algorithmic}
\label{alg:generate_ts}
\end{algorithm}


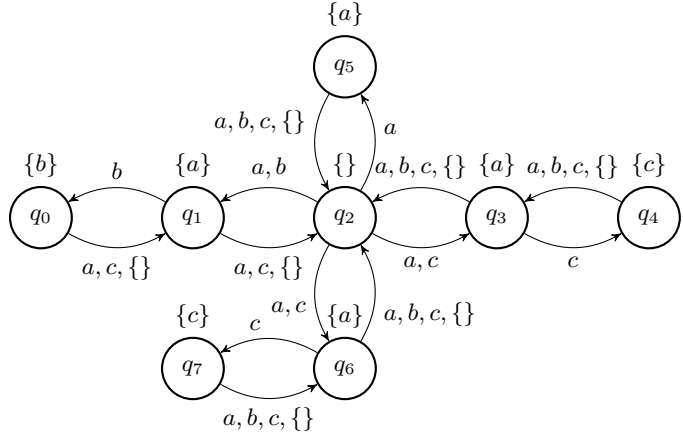
\begin{figure}[h]
\centering
 \resizebox{1.0\linewidth}{!}{
\begin{tikzpicture}
\node[state,label={above:$\lbrace b\rbrace$}] (q0) {$q_0$};
        \node[state,right of=q0,label={above:$\lbrace a\rbrace$}] (q1) {$q_1$};
        \node[state,right of=q1,label={above:$\lbrace\rbrace$}] (q2) {$q_2$};
        \node[state,right of=q2,label={above:$\lbrace a\rbrace$}] (q3) {$q_3$};
        \node[state,right of=q3,label={above:$\lbrace c\rbrace$}] (q4) {$q_4$};

        \node[state,above of=q2,label={above:$\lbrace a\rbrace$}] (q5) {$q_5$};
        
        \node[state,below of=q2,label={above:$\lbrace a\rbrace$}] (q6) {$q_6$};
        \node[state,left of=q6,label={above:$\lbrace c\rbrace$}] (q7) {$q_7$};
        
        \draw 
        (q0) edge[bend right, below] node{$a,c,\{\}$} (q1)
        (q1) edge[bend right, below] node{$a,c,\{\}$} (q2)
        (q2) edge[bend right, below] node{$a,c$} (q3)
        (q3) edge[bend right, below] node{$c$} (q4)
        
        (q1) edge[bend right, above] node{$b$} (q0)
        (q2) edge[bend right, above] node{$a,b$} (q1)
        (q3) edge[bend right, above] node{$a,b,c,\{\}$} (q2)
        (q4) edge[bend right, above] node{$a,b,c,\{\}$} (q3)
        
        (q6) edge[bend right, below right] node{$a,b,c,\{\}$} (q2)
        (q7) edge[bend right, below] node{$a,b,c,\{\}$} (q6)
        (q2) edge[bend right, below left] node{$a,c$} (q6)
        (q6) edge[bend right, above left] node{$c$} (q7)

        (q5) edge[bend right, above left] node{$a,b,c,\{\}$} (q2)
        (q2) edge[bend right, above right] node{$a$} (q5);
\end{tikzpicture}
}
\caption{Unpruned TS created by Alg.~\ref{alg:generate_ts} on the environment in Fig.~\ref{fig:env}. State labels appear above each state. Transition labels correspond to task policies that enable a transition.}
\label{unpruned_ts}
\end{figure}

\par
A TS constructed in this manner could be non-deterministic. For example, state $q_2$ in Fig.~\ref{unpruned_ts} has multiple outgoing transitions labeled $a$. 
To resolve this non-determinism, we propose a method for pruning the TS.
To prune, we will remove transitions and policies in transition labels that introduce non-determinism, by checking for the specific cases of: 1) Equivalency; 2) Ambiguity; and 3) Feasibility; with the methods for mitigating these cases later described in Sec.~\ref{sec:prune_ts}. 

\begin{remark}
We note that our approach differs from standard approaches to determinising a system, such as subset construction~\cite{hopcroft2001introduction}. Those methods track the multiple states that could be reached under a given action. We wish to find a representation such that executing a policy corresponds exactly with transitioning from one region to another.
\end{remark}

\subsection{Transition System Pruning}
\label{sec:prune_ts}

When we follow the procedure outlined in Sec.~\ref{sec:generate_ts}, we capture how states are connected, but the resulting TS state and transition labels can introduce non-determinism.
To mitigate such problems, we introduce a TS pruning method, which removes symbols from transition labels. Algorithm~\ref{alg:prune} shows the overall pruning procedure. Each \texttt{case} function in the algorithm is explained below, along with an explanation of the results of pruning the TS from Figure~\ref{unpruned_ts}.

\begin{algorithm}[h]
  \caption{Transition System Prune}
  \begin{algorithmic}[1]
\Procedure{Prune}{$S, T, \Sigma$}

\State $S, T \gets  \texttt{case1}(S, T, \Sigma)$
\For{$s \in S$}
    \State$T \gets  \texttt{case2}(s, T, \Sigma)$
    \State$T \gets  \texttt{case3}(s, T, \Sigma)$
\EndFor
\State$T \gets  \texttt{emptyCleanup}(T)$
\State \Return $S, T$ 
\EndProcedure
\end{algorithmic}
\label{alg:prune}
\end{algorithm}

\subsubsection{Case 1: Equivalent States}
\label{sec:case1}

\par
In a TS where multiple branches from a parent state contain the same state and transition labels, taking a policy from one of the symbols shared on the transition label could take the agent to any of the duplicate child state regions. 
Since there is not meaningful distinction between the two for the purposes of satisfying a specification,
they are effectively equivalent. Therefore, all duplicate branches are merged into one to simplify the TS.

\textit{\textbf{Case 1:} If there are duplicate states, where the state label, outgoing transition labels, and incoming transition labels are the same between states, then combine the states into one.}

\par
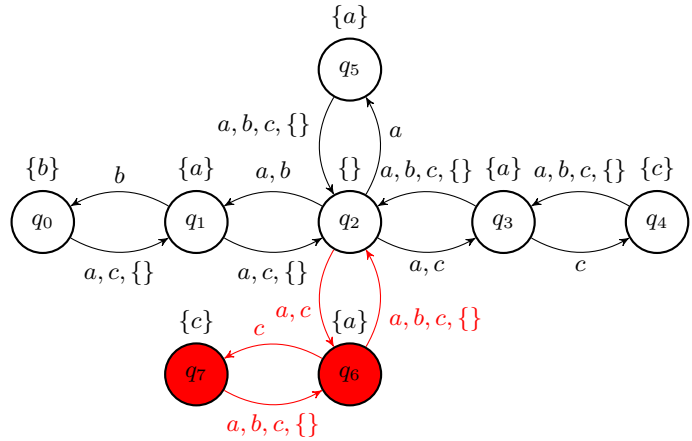
\begin{figure}[h]
  \centering
    \resizebox{1.0\linewidth}{!}{
    \begin{tikzpicture}
    \node[state,label={above:$\lbrace b\rbrace$}] (q0) {$q_0$};
        \node[state,right of=q0,label={above:$\lbrace a\rbrace$}] (q1) {$q_1$};
        \node[state,right of=q1,label={above:$\lbrace\rbrace$}] (q2) {$q_2$};
        \node[state,right of=q2,label={above:$\lbrace a\rbrace$}] (q3) {$q_3$};
        \node[state,right of=q3,label={above:$\lbrace c\rbrace$}] (q4) {$q_4$};

        \node[state,above of=q2,label={above:$\lbrace a\rbrace$}] (q5) {$q_5$};
        
        \node[state,below of=q2,label={above:$\lbrace a\rbrace$},fill={rgb:red,1}] (q6) {$q_6$};
        \node[state,left of=q6,label={above:$\lbrace c\rbrace$},fill={rgb:red,1}] (q7) {$q_7$};
        
        \draw 
        (q0) edge[bend right, below] node{$a,c,\{\}$} (q1)
        (q1) edge[bend right, below] node{$a,c,\{\}$} (q2)
        (q2) edge[bend right, below] node{$a,c$} (q3)
        (q3) edge[bend right, below] node{$c$} (q4)
        
        (q1) edge[bend right, above] node{$b$} (q0)
        (q2) edge[bend right, above] node{$a,b$} (q1)
        (q3) edge[bend right, above] node{$a,b,c,\{\}$} (q2)
        (q4) edge[bend right, above] node{$a,b,c,\{\}$} (q3)
        
        (q6) edge[bend right, below right, color=red] node{$a,b,c,\{\}$} (q2)
        (q7) edge[bend right, below, color=red] node{$a,b,c,\{\}$} (q6)
        (q2) edge[bend right, below left, color=red] node{$a,c$} (q6)
        (q6) edge[bend right, above left, color=red] node{$c$} (q7)

        (q5) edge[bend right, above left] node{$a,b,c,\{\}$} (q2)
        (q2) edge[bend right, above right] node{$a$} (q5);
    \end{tikzpicture}
    }
\caption{State labels appear above each state. Transition labels correspond to task policies that enable a transition. Red states and transitions are removed via \texttt{case1} to create Figure \ref{case2}. Symbols \texttt{a, b, c, \{\}} are deleted to remove equivalency in the system.}
\label{case1}
\end{figure}

 
Figure~\ref{case1} highlights the changes in the TS after \texttt{case1} in Alg.~\ref{alg:prune} is executed. Algorithm \texttt{case1} identifies that $q_2$ has branches that are equivalent. The two equivalent branches are 1) the branch containing $q_3$ and $q_4$ and 2) the branch containing q6 and $q_7$. 
In Fig.~\ref{case1}, the two equivalent branches get combined into the branch containing $q_3$ and $q_4$.

\subsubsection{Case 2: Ambiguous Transitions}
\label{sec:case2} 

\par
If a state has multiple transition labels that contain the same symbol, it is uncertain which transition will be followed when the corresponding policy is executed. 
Because we seek a method that is zero-shot, we perform no additional checks or training on the policy to see how it would behave if run in the state region; therefore, we wish to keep at most once outgoing transition labeled with that symbol.

\textit{\textbf{Case 2:} If any outgoing transitions from a state share a symbol in the transition label, only keep the symbol in the transition with the least distance to the state labeled with the shared symbol, according to MV semantics. If all the distances to the state that is labeled with the shared symbol are the same, remove the symbol from all the transition labels of the state.}

\par
Algorithm~\ref{alg:case2} shows the procedure for \texttt{case2}. 
For a given state, the algorithm finds the set of outgoing transitions labeled by a given symbol $\sigma$ (lines 2--3). If there is more than one such transition, determines the state(s) with the largest distance according to MV semantics (lines 5--7) and removes $\sigma$ from the corresponding transitions. The process repeats until there is at most one transition with label $\sigma$.

\begin{algorithm}[h]
\caption{Case 2 Prune}
\begin{algorithmic}[1]
\Procedure{case2}{$s,T,\Sigma$}
\For{$\sigma \in \Sigma$}
    \State $t_\sigma \gets \lbrace t \in T \mid \sigma \in L(t) \rbrace$
    \While{$\vert t_\sigma \vert \geq 2$}
        \State $s_{\sigma} \gets \lbrace s \in S \mid \sigma \in L(s)\rbrace$
        \State $d_t \gets max_{(t, s_{\sigma})}(d(s, s_{\sigma}))$
        \State $t_{far} \gets \lbrace t \in t_{\sigma} \mid d_t \texttt{ is greatest} \rbrace$
        \State {\texttt{delete $\sigma$ from $t_{far}$}}
        \State {\texttt{delete $t_{far}$ from $t_\sigma$}}
    \EndWhile
\EndFor
\Return $T$
\EndProcedure
\end{algorithmic}
\label{alg:case2}
\end{algorithm}

\par
Figure \ref{case2} highlights the changes in the TS after \texttt{case2} is executed. No labels $a$ are kept on outgoing transitions from $q_2$, because MV semantics cannot distinguish them. The label $c$ is removed from the transition linking $q_3$ to $q_2$, because MV semantics will result in an agent transitioning from $q_3$ to $q_4$ under a policy associated with task $c$. 

\begin{figure}[h] 
\centering
\resizebox{1.0\linewidth}{!}{
\begin{tikzpicture}
\node[state,label={above:$\lbrace b\rbrace$}] (q0) {$q_0$};
    \node[state,right of=q0,label={above:$\lbrace a\rbrace$}] (q1) {$q_1$};
    \node[state,right of=q1,label={above:$\lbrace\rbrace$}] (q2) {$q_2$};
    \node[state,right of=q2,label={above:$\lbrace a\rbrace$}] (q3) {$q_3$};
    \node[state,right of=q3,label={above:$\lbrace c\rbrace$}] (q4) {$q_4$};
    \node[state,above of=q2,label={above:$\lbrace a\rbrace$}] (q5) {$q_5$};
    
    \draw 
    (q0) edge[bend right, below] node{$a,c,\{\}$} (q1)
    (q1) edge[bend right, below] node{$a,c,\{\}$} (q2)
    (q2) edge[bend right, below] node{$\textcolor{red}{a},c$} (q3)
    (q3) edge[bend right, below] node{$c$} (q4)
    
    (q1) edge[bend right, above] node{$b$} (q0)
    (q2) edge[bend right, above] node{$\textcolor{red}{a},b$} (q1)
    (q3) edge[bend right, above] node{$a,b,\textcolor{red}{c},\{\}$} (q2)
    (q4) edge[bend right, above] node{$a,b,c,\{\}$} (q3)
    
    (q5) edge[bend right, above left] node{$a,b,c,\{\}$} (q2)
    (q2) edge[bend right, above right] node{$\textcolor{red}{a}$} (q5);
\end{tikzpicture}
}
\caption{State labels appear above each state. Transition labels correspond to task policies that enable a transition. Red label symbols are removed via \texttt{case2} to create Figure \ref{case3}. Symbols \texttt{a} and \texttt{c} are deleted to remove unambiguous transition labels from the system.}
\label{case2}
\end{figure}

\subsubsection{Case 3: Ineffectual Transitions and Feasibility}
\label{sec:case3}

\par
This case only arises when there are multiple states containing the same symbol label during the initial TS creation (Alg.~\ref{alg:generate_ts}). Each duplicate state will have an outgoing transition label containing the same symbol as its own label, to get to the other states that share the same symbol label. 
We prune the symbol from the outgoing transition labels as running the policy for generating a symbol while already in the region that produces the symbol will not cause the agent to transition out of its current state. Therefore, since the state does not change, the symbol on the label is ineffectual. 

\textit{\textbf{Case 3:} If a state shares the same label as any outgoing transition, remove the label from those transitions.}

\par
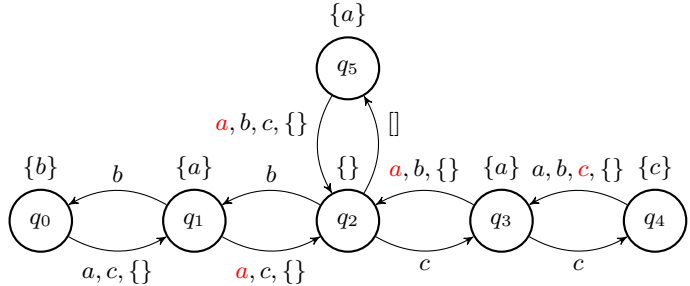
\begin{figure}[h]
  \centering
    \resizebox{1.0\linewidth}{!}{
    \begin{tikzpicture}
        \node[state,label={above:$\lbrace b\rbrace$}] (q0) {$q_0$};
        \node[state,right of=q0,label={above:$\lbrace a\rbrace$}] (q1) {$q_1$};
        \node[state,right of=q1,label={above:$\lbrace\rbrace$}] (q2) {$q_2$};
        \node[state,right of=q2,label={above:$\lbrace a\rbrace$}] (q3) {$q_3$};
        \node[state,right of=q3,label={above:$\lbrace c\rbrace$}] (q4) {$q_4$};
        \node[state,above of=q2,label={above:$\lbrace a\rbrace$}] (q5) {$q_5$};
        
        \draw 
        (q0) edge[bend right, below] node{$a,c,\{\}$} (q1)
        (q1) edge[bend right, below] node{$\textcolor{red}{a},c,\{\}$} (q2)
        (q2) edge[bend right, below] node{$c$} (q3)
        (q3) edge[bend right, below] node{$c$} (q4)
        
        (q1) edge[bend right, above] node{$b$} (q0)
        (q2) edge[bend right, above] node{$b$} (q1)
        (q3) edge[bend right, above] node{$\textcolor{red}{a},b,\{\}$} (q2)
        (q4) edge[bend right, above] node{$a,b,\textcolor{red}{c},\{\}$} (q3)
        
        (q5) edge[bend right, above left] node{$\textcolor{red}{a},b,c,\{\}$} (q2)
        (q2) edge[bend right, above right] node{$[]$} (q5);
    \end{tikzpicture}
    }
\caption{State labels appear above each state. Transition labels correspond to task policies that enable a transition. Red label symbols are removed via \texttt{case3} to create Fig.~\ref{emptyCleanup}. Symbols $a$ and $c$ are deleted to remove system infeasibility from the system.}
\label{case3}
\end{figure}

Figure \ref{case3} highlights the changes in the TS after \texttt{case3} in Alg.~\ref{alg:prune} is executed.
State $q_1$'s label is $a$, and the transition from $q_1$ to $q_2$ contains $a$, so $a$ is removed from that transition. The same logic applies to the other highlighted labels.

\subsubsection{Empty Cleanup}

\begin{figure}
  \centering
    \resizebox{1.0\linewidth}{!}{
    \begin{tikzpicture}
        \node[state,label={above:$\lbrace b\rbrace$}] (q0) {$q_0$};
        \node[state,right of=q0,label={above:$\lbrace a\rbrace$}] (q1) {$q_1$};
        \node[state,right of=q1,label={above:$\lbrace\rbrace$}] (q2) {$q_2$};
        \node[state,right of=q2,label={above:$\lbrace a\rbrace$}] (q3) {$q_3$};
        \node[state,right of=q3,label={above:$\lbrace c\rbrace$}] (q4) {$q_4$};
        \node[state,above of=q2,label={above:$\lbrace a\rbrace$}] (q5) {$q_5$};
        
        \draw 
        (q0) edge[bend right, below] node{$a,c,\textcolor{red}{\{\}}$} (q1)
        (q1) edge[bend right, below] node{$c,\textcolor{red}{\{\}}$} (q2)
        (q2) edge[bend right, below] node{$c$} (q3)
        (q3) edge[bend right, below] node{$c$} (q4)
        
        (q1) edge[bend right, above] node{$b$} (q0)
        (q2) edge[bend right, above] node{$b$} (q1)
        (q3) edge[bend right, above] node{$b,\textcolor{red}{\{\}}$} (q2)
        (q4) edge[bend right, above] node{$a,b,\textcolor{red}{\{\}}$} (q3)
        
        (q5) edge[bend right, above left] node{$b,c,\textcolor{red}{\{\}}$} (q2)
        (q2) edge[bend right, above right, red] node{$[]$} (q5);
    \end{tikzpicture}
    }
\caption{State labels appear above each state. Transition labels correspond to task policies that enable a transition. Red label symbols are removed via \texttt{emptyCleanup}.
}
\label{emptyCleanup}
\end{figure}
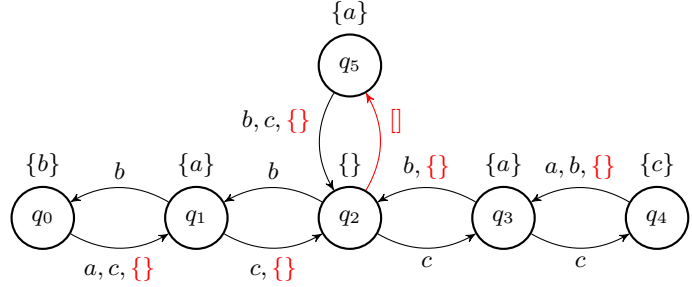

We require final cleanup after the TS has been generated through the 3 cases. Since there will never be a policy for spaces that produce no symbols, the policy $\emptyset$ is removed from all transition labels. Also, all transitions with no transition label are removed, as there is no policy that can take the agent between the connected states. The transition will create uncertainty when the product is taken with the TS, so it is imperative it is removed.
Figure \ref{emptyCleanup} highlights the changes in the TS after \texttt{emptyCleanup} in Alg.~\ref{alg:prune} is executed. 

\subsection{Product between Transition System and B\"{u}chi Automaton}

Given a fully pruned TS with labels from $\Sigma$, we create a B\"{u}chi automaton using a an LTL formula specification $\phi$ over $\Sigma$. We can then construct a Cartesian product between the TS and automaton, preserving the transition labels from the TS. The resulting product automaton (PA) can be used with typical sequence generation methods to find a satisfying sequence~\cite{beltabook2017}.

\subsection{Theoretical Analysis}
In this section, we propose three theorems about our method.

\begin{theorem}\label{thm:deterministic}
    The resulting pruned TS from Sec.~\ref{sec:prune_ts} is deterministic.
\end{theorem}

\begin{sketch}
We note two ways in which a TS may be non-deterministic. First, 
a TS may transition to
multiple states given a single action (either indistinguishable from each other or not), which cases 1 and 2 address. 
Second, given a state and an action, the TS may stay in the same state or transition to another state. The MV semantics of our policies preclude moving in favor of self-loops, which case 3 addresses.
\end{sketch}

\begin{theorem}\label{thm:realizable}
    The resulting pruned TS from Sec.~\ref{sec:prune_ts} contains no unrealizable transitions.
\end{theorem}

\begin{sketch}
The naive TS construction captures all potential transitions that an RL policy enforces. Our pruning process respects the MV semantics as defined in \ref{sec:background}. By pruning transitions to farther states with same label, we prune states that would never be reached under MV semantics, because those semantics prioritize producing fewer symbols.
\end{sketch}

\begin{theorem}
    Satisfying an LTL specification using product construction with our pruned TS is sound.
\end{theorem}

\begin{sketch}
    This follows directly from Theorems~\ref{thm:deterministic} and~\ref{thm:realizable}. Those theorems imply that the agent executing its RL policies is a simulation of the pruned TS, and the usual guarantees on the PA hold; therefore finding a satisfying $\tau$ in the PA using our TS is sound.
\end{sketch}

\section{Results}
\subsection{Simulation}

To demonstrate our logic, we used a high-dimensional video game environment~\cite{tasse2020booleantaskalgebrareinforcement}. This is a grid-world environment with 6 possible items: every combination of colors beige, blue, and purple, with shapes circle and square. We consider LTL formulas over the set propositions \{\textcolor{brown}{w (white)}, \textcolor{blue}{b (blue)}, \textcolor{violet}{p (purple)}, $\CIRCLE$ (circle), $\blacksquare$ (square)\}. These traits can be composed in a Boolean fashion, e.g., $\textcolor{blue}{\blacksquare}:=\textcolor{blue}{b}\wedge\blacksquare$. Each grid cell is either unoccupied, represented by empty set $\emptyset$, or contains an object characterized by a shape and a color from the set propositions.

Our simulations are executed in Python 3.7. Our TS and PA are constructed using NetworkX~\cite{hagberg2008} and a modified version of  LOMAP~\footnote{\url{https://github.com/wasserfeder/lomap}}.
The policies used in this experiment are trained using the RL methodology from~\citet{leahy2023safetyaware}. All policies are trained using MV semantics.
During training the environment randomly spawns items and the player's start position. The observation space is down sampled 84x84 RGB images of the world and the action space is the 4 cardinal directions: up, down, left right. See~\citet{leahy2023safetyaware} for more details on training. 

One policy is trained for each of the six primitive tasks above.
The composition of policies is performed zero-shot via the method of~\citet{leahy2023safetyaware}.

\subsection{Case Study}

\par
For the demonstration, we show that two different LTL formulas of varying complexities with two different map configurations produce the expected symbols. The same policies are used for both demonstrations.

\begin{figure}
  \centering
  \begin{subfigure}[b]{.5\linewidth}
    \centering
    \resizebox{\linewidth}{!}{
    \begin{tikzpicture}
        \node[state,label={above:$\lbrace \textcolor{brown}{\CIRCLE} \rbrace$}] (q0) {$q_0$};
        \node[state,below right of=q0,label={above:$\lbrace \textcolor{violet}{\CIRCLE} \rbrace$}] (q1) {$q_1$};
        \node[state,below right of=q1, label={above:$\lbrace \textcolor{blue}{\CIRCLE} \rbrace$}] (q2) {$q_2$};
        \node[state,below right of=q2,label={above:$\lbrace \rbrace$}] (q4) {$q_4$};
        \node[state, right of = q1, above of=q4, label={above:$\lbrace \textcolor{blue}{\blacksquare} \rbrace$}] (q3) {$q_3$};

        \draw
        (q4) edge[out=0,in=-90] node{$\textcolor{blue}{\blacksquare}$}(q3)
        (q4) edge[out=180,in=-90] node{$\textcolor{blue}{\CIRCLE}$} (q2)
        (q4) edge[out=180,in=-90] node{$\textcolor{violet}{\CIRCLE}$} (q1)
        (q4) edge[out=180,in=-90] node{$\textcolor{brown}{\CIRCLE}$} (q0)

        (q3) edge[out=-135,in=45] node{$\textcolor{blue}{\CIRCLE}$} (q2)
        (q3) edge[out=180,in=45] node{$\textcolor{violet}{\CIRCLE}$} (q1)
        (q3) edge[out=135,in=45] node{$\textcolor{brown}{\CIRCLE}$} (q0)

        (q2) edge[out=0,in=-112.5] node{$\textcolor{blue}{\blacksquare} $} (q3)
        (q1) edge[out=0,in=202.5] node{$\textcolor{blue}{\blacksquare} $} (q3)
        (q0) edge[out=0,in=157.5] node{$\textcolor{blue}{\blacksquare} $} (q3);
        



\end{tikzpicture}}
    \caption{Generated Transition System from Map.}
    \label{ts_state1}
  \end{subfigure}%
  \hfill
  \begin{subfigure}[b]{.4\linewidth}
    \centering
    \resizebox{\linewidth}{!}{
    \includegraphics[width=.48\linewidth]{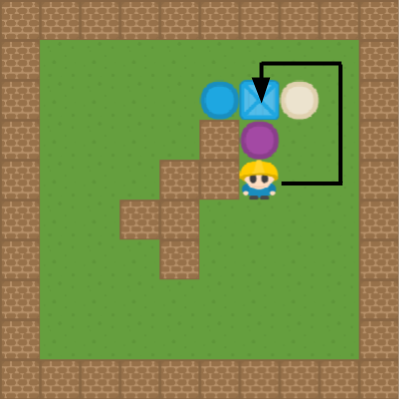}
    }
    \caption{Path Agent Takes Executing Policy $\pi_{\textcolor{blue}{\blacksquare}}$ from PA.}
    \label{agent_path_state_1_F_e}
  \end{subfigure}
\caption{Pipeline for $\LTLeventually \blacksquare$.}
\label{multifig:starte_state_1_LTL_F_e}
\end{figure}

The first example is shown in Fig.~\ref{multifig:starte_state_1_LTL_F_e}. This example uses the simple LTL specification $\LTLeventually \blacksquare$, which translates to ``eventually square". Our logic produces the shortest word [\textcolor{blue}{$\blacksquare$}], which translates to the Boolean composition policy $\pi_{\textcolor{blue}{\blacksquare}}:=\pi_{\textcolor{blue}{b}} \wedge \pi_{\blacksquare}$.
Since the agent is trained using minimum violation, Fig.~\ref{agent_path_state_1_F_e} shows that the agent following policy $\pi_{\textcolor{blue}{\blacksquare}}$ does not enter any area containing another color or shape. The path is optimal following our logic as there are not any additional symbols encountered along the path and the path never violates the LTL specification.

\begin{figure}
  \centering
  \begin{subfigure}[b]{.5\linewidth}
    \centering
    \resizebox{\linewidth}{!}{
        \begin{tikzpicture}


        \node[state,label={above:$\lbrace \textcolor{violet}{\blacksquare} \rbrace$}] (q0) {$q_0$};
        \node[state,below right of=q0,label={above:$\lbrace \textcolor{brown}{\CIRCLE} \rbrace$}] (q1) {$q_1$};
        \node[state,below right of=q1, label={above:$\lbrace \textcolor{violet}{\CIRCLE} \rbrace$}] (q2) {$q_2$};
        \node[state,below right of=q2, label={above:$\lbrace \textcolor{blue}{\CIRCLE} \rbrace$}] (q3) {$q_3$};
        \node[state,below right of=q3,label={above:$\lbrace \rbrace$}] (q5) {$q_5$};
        \node[state, above right of = q2,label={above:$\lbrace \textcolor{blue}{\blacksquare} \rbrace$}] (q4) {$q_4$};

        \draw
        (q5) edge[out=0,in=0] node{$\textcolor{blue}{\blacksquare} $} (q4)
        (q5) edge[out=180,in=-90] node{$\textcolor{blue}{\CIRCLE}$}(q3)
        (q5) edge[out=180,in=-90] node{$\textcolor{violet}{\CIRCLE}$} (q2)
        (q5) edge[out=180,in=-90] node{$\textcolor{brown}{\CIRCLE}$} (q1)
        (q5) edge[out=180,in=-90] node{$\textcolor{violet}{\blacksquare}$} (q0)

        (q4) edge[out=-90,in=45] node{$\textcolor{blue}{\CIRCLE}$} (q3)
        (q4) edge[out=-135,in=45] node{$\textcolor{violet}{\CIRCLE}$} (q2)
        (q4) edge[out=180,in=45] node{$\textcolor{brown}{\CIRCLE}$} (q1)
        (q4) edge[out=135,in=45] node{$\textcolor{violet}{\blacksquare}$} (q0)

        (q3) edge[out=0,in=-67.5] node{$\textcolor{blue}{\blacksquare} $} (q4)
        (q2) edge[out=0,in=-112.5] node{$\textcolor{blue}{\blacksquare} $} (q4)
        (q1) edge[out=0,in=202.5] node{$\textcolor{blue}{\blacksquare} $} (q4)
        (q0) edge[out=0,in=157.5] node{$\textcolor{blue}{\blacksquare} $} (q4);
        



    \end{tikzpicture}}
    \caption{Generated Transition System from Map.}
    \label{ts_state3}
  \end{subfigure}%
  \hfill
  \begin{subfigure}[b]{.4\linewidth}
    \centering
    \resizebox{\linewidth}{!}{
    \includegraphics[width=.48\linewidth]{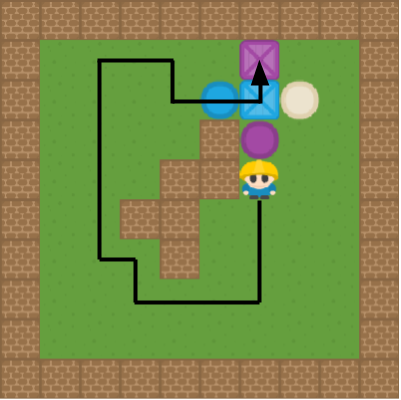}
    }
    \caption{Path Agent Takes Executing Policies [$\pi_{\textcolor{blue}{\CIRCLE}}, \pi_{\textcolor{blue}{\blacksquare}}, \pi_{\textcolor{violet}{\blacksquare}}$] from PA.}
    \label{agent_path_state_3_F_(b_&_!_e)_&_F_c}
  \end{subfigure}
\caption{Pipeline for $\LTLeventually (\textcolor{blue}{b} \wedge \Not \blacksquare) \wedge \LTLeventually \textcolor{violet}{p}$.}
\label{multifig:starte_state_3_LTL_F_(b_&_!_e)_&_F_c}
\end{figure}

The second example is shown in Figure \ref{multifig:starte_state_3_LTL_F_(b_&_!_e)_&_F_c}. This example uses the more complex LTL specification  $\LTLeventually (\textcolor{blue}{b} \wedge \Not \blacksquare) \wedge \LTLeventually \textcolor{violet}{p}$, which translates to ``eventually (blue and not square) and eventually purple". Our logic produces the word 
[$\textcolor{blue}{\CIRCLE}, \textcolor{blue}{\blacksquare}, \textcolor{violet}{\blacksquare}$], which corresponds to the sequence of Boolean composition policies given by [$\pi_{\textcolor{blue}{\CIRCLE}}, \pi_{\textcolor{blue}{\blacksquare}}, \pi_{\textcolor{violet}{\blacksquare}}$]:=[$\pi_{\textcolor{blue}{b}} \wedge \pi_{\CIRCLE}, \pi_{\textcolor{blue}{b}} \wedge \pi_{\blacksquare}, \pi_{\textcolor{violet}{p}} \wedge \pi_{\blacksquare}$]. The agent progresses along the list of policies in the order provided, so first the agent begins executing $\pi_{\textcolor{blue}{\CIRCLE}}$. Once the agent has reached a region that produces $\textcolor{blue}{b} \LTLand \CIRCLE$, the agent transitions to executing the next policy. The agent is done when it has reached a region that produces the symbols of the final policy. In this instance, the agent is finished when it enters the region that produces $\textcolor{violet}{p} \LTLand \blacksquare$.

Since the agent is trained using MV semantics, Fig.~\ref{agent_path_state_3_F_(b_&_!_e)_&_F_c} shows that the agent following a policy does not enter any cell containing another color or shape. That is why the first path to the blue circle takes the agent the long way around the center obstacle, as it will not travel the shorter path to blue circle and encounter additional symbols from the other region; therefore the path is optimal following our logic as there are not any additional symbols encountered along the path and the path never violates the LTL specification.

\par
\begin{remark} A trade-off of our approach is demonstrated in this case study. The agent does not take the shortest path in the environment, $\{\textcolor{violet}{\CIRCLE}, \textcolor{blue}{\blacksquare}, \textcolor{blue}{\CIRCLE}\}$, since we only consider path length in the automaton.
The paths $\{\textcolor{violet}{\CIRCLE}, \textcolor{blue}{\blacksquare}, \textcolor{blue}{\CIRCLE}\}$ and $\{\textcolor{blue}{\CIRCLE}, \textcolor{blue}{\blacksquare}, \textcolor{violet}{\blacksquare}\}$ both have automaton path length 3.
This is one of the primary trade-offs for zero-shot satisfaction, and methods such as RM can use fine-tuning to address this trade-off, but require additional training episodes.\end{remark}

\subsection{Comparison}
Our main contribution is an approach that is safety aware and zero-shot. We compare our approach, \ourMethod, to two other state-of-the-art approaches BC and RM. \ourMethod trains tasks primitives using safety properties before run time and combines models temporally as needed using composition at run time (zero-shot) using environmental information to satisfy the specification. These safety-focused policies are MV policies. \booleanComposition~\cite{tasse2020booleantaskalgebrareinforcement} trains task primitives before run time and combines models as needed using composition. \rewardMachines~\cite{icarte2018} trains a task policy at run time based on a specification provided at run time. 

To demonstrate the necessity of safety primitive policies, we train primitive policies using \booleanComposition and replace our safety primitive policies in our pipeline with their primitive policies. To demonstrate the run time benefits of zero-shot composition, we implemented \rewardMachines on the same Boxman environment, trained using the default/recommended hyperparameters, and directly compare our entire pipeline's performance to \rewardMachines's performance.  

We compare the approaches based on three metrics 1) path safety; 2) training time; and 3) specification processing time. Path safety ensures that when a primitive policy, or composition of primitive policies, is being executed, no other symbol is produced unless necessary. Training time is the time for a primitive policy to be fully trained. 
This is not applicable for \rewardMachines as there are no primitive policies to train. Specification processing time is the time for the approach to recalculate the approach based on a new LTL specification. We provide \rewardMachines with the new LTL specification as a state machine which they train on, and we create the state machine from a translation of the specification automation created by SPOT~\cite{spot}. The MV and \booleanComposition approaches are provided a new LTL specification as a string. 


\begin{table}
\centering
\begin{tabular}{ |p{2cm}||p{2.5cm}|p{2.5cm}| }
 \hline
 \multicolumn{3}{|c|}{Primitive Policy Training Time} \\
 \hline
Policy Type & MV Time (s) & \booleanComposition Time (s) \\
 \hline
 Blue & 270,602 &  224,564 \\
 \hline
 White & 272,423 & 192,312 \\
 \hline
 Purple & 321,445 & 237,616 \\
 \hline
\end{tabular}
\caption{Time to train primitive models for object colors.}
\label{tab:time_train_tasks}
\end{table}

\begin{table}
\centering
\begin{tabular}{ |p{3.55cm}||p{1.75cm}|p{1.75cm}|}
 \hline
 \multicolumn{3}{|c|}{Execution Time for a New LTL Specification} \\
 \hline
 LTL Specification & \ourMethod Time (s) & \rewardMachines (s)  \\
 \hline
 $\LTLeventually (\textcolor{blue}{b} \&  \blacksquare)$ & 0.0371 & 19151 
 \\
 \hline
 $\LTLeventually (\textcolor{purple}{p} \& \CIRCLE)$ & 0.0422 &  30647 
 \\
 \hline
 $\LTLeventually \textcolor{blue}{b} \& \LTLalways \Not \blacksquare$ & 0.0552 & 1336 
 \\
 \hline
 $\LTLalways(\LTLeventually(\textcolor{blue}{b} \&  \blacksquare)) \& \LTLalways(\LTLeventually(\textcolor{purple}{p} \& \CIRCLE))$ &  0.0305 & 3959 \\  
 \hline
\end{tabular}
\caption{Time to reprocess given a new LTL specification.}
\label{tab:execution_time}
\end{table}

\begin{table}
\centering
\begin{tabular}{ |p{1.85cm}||p{1.6cm}|p{1.93cm}|p{1.4cm}|}
 \hline
 \multicolumn{4}{|c|}{Number of ``Unsafe" Symbols Collected} \\
 \hline
 LTL Specification & \ourMethod & \ourMethod + \booleanComposition Policies & \rewardMachines  \\
 \hline
 $\LTLeventually (\textcolor{blue}{b} \&  \blacksquare)$ & \textcolor{green}{0} &  \textcolor{green}{1} & \textcolor{green}{1} \\
 \hline
 $\LTLeventually (\textcolor{purple}{p} \& \CIRCLE)$   & \textcolor{green}{0} & \textcolor{green}{0} & \textcolor{green}{0} \\
 \hline
 $\LTLeventually \textcolor{blue}{b} \& \LTLalways \Not \blacksquare$  & \textcolor{green}{0} & \textcolor{red}{0} & \textcolor{red}{2} \\
 \hline
  $\LTLalways(\LTLeventually(\textcolor{blue}{b} \&  \blacksquare))$ $\& \LTLalways(\LTLeventually(\textcolor{purple}{p} \& \CIRCLE))$ & \textcolor{green}{0} & \textcolor{green}{0} & \textcolor{red}{0} \\
 \hline
 \hhline{|=|=|=|=|}
 Total & 0 & 1 & 3 \\
 \hline
\end{tabular}
\caption{Number of symbols collected that are not specified in the policy or specification. Green indicates the specification was satisfied by the the trajectory (agent behavior), and red indicates the specification was not satisfied by the trajectory. \ourMethod with \booleanComposition policies is our framework with our safety policies swapped for \booleanComposition policies.}
\label{tab:unsafe_symbol_collection}
\end{table}

Table \ref{tab:time_train_tasks} shows that MV primitive policies take longer to train than non-MV policies. \ourMethod takes, on average, 32\% longer than \booleanComposition to train primitive policies. 

Table \ref{tab:execution_time} shows that upon a change in the LTL specification, our mechanism takes significantly less time to reprocess, as we are not training whereas \rewardMachines requires retraining. For a new LTL specification, our reward is, on average, 99.99\% quicker than \rewardMachines. We observe that reward machines that include a transition that uses $\LTLor$ take longer to train than reward machines without $\LTLor$. The reward machines for specifications $\LTLeventually (\textcolor{blue}{b} \&  \blacksquare)$ and $\LTLeventually (\textcolor{purple}{p} \& \CIRCLE)$ include $\LTLor$, so they take on average 8.4\% longer to train than the specifications  $\LTLeventually \textcolor{blue}{b} \& \LTLalways \Not \blacksquare$ and
 $\LTLalways(\LTLeventually(\textcolor{blue}{b} \&  \blacksquare)) \& \LTLalways(\LTLeventually(\textcolor{purple}{p} \& \CIRCLE))$, which do not include $\LTLor$ in their reward machines.

Table \ref{tab:unsafe_symbol_collection} shows that \ourMethod's additional training for safety results in no additional symbols being generated other than the symbol for the primitive policy and we are the only approach to consistently satisfy the specification. 
Table \ref{tab:unsafe_symbol_collection} also shows that other approaches produce extraneous symbols. 
For the specification $\LTLeventually \textcolor{blue}{b} \& \LTLalways \Not \blacksquare$, \ourMethod with \booleanComposition policies only collects the purple circle symbol, it does not reach any blue symbols, so it fails the specification even though it doesn't produce any unsafe behavior. For the same specification \rewardMachines collects purple circle and blue square, but square is explicitly not allowed, so it fails the specification. Finally, for the specification $\LTLalways(\LTLeventually(\textcolor{blue}{b} \&  \blacksquare)) \& \LTLalways(\LTLeventually(\textcolor{purple}{p} \& \CIRCLE))$, \rewardMachines does not converge as the reward machine never collects any rewards -- we note that we used the default recommended training parameters and with more parameter tuning or a different reward machine representation of the LTL specification the policy may have converged.

Our results show that our zero-shot approach requires no additional training per specification, and the paths our approach produces are safe and feasible.

\section{Conclusion}

\par
We present \ourMethod, an end-to-end zero-shot approach for executing an LTL task specification. To encode environment topology, we create a TS representation of the environment. 
Creating the TS introduces non-determinism and ambiguity, which we resolve via pruning. Our pruned TS is deterministic, contains only feasible transitions, and is sound. We create a B\"{u}chi automaton from an LTL specification then take the product between the TS and B\"{u}chi automaton to create a PA, which encodes both environment topology and the task specification. We use the PA to find a satisfying path for the agent to reach an acceptance state in the LTL specification. This satisfying path is a list of MV policies for the agent to execute throughout its environment traversal. 
\ourMethod is validated via simulation.
We also show that our MV policies do not produce any extra symbols, unless necessary, and verify that the policies produce the expected symbols. We also compare processing and training time to other state of the art approaches, showing that our approach is safer and more adaptable.

\par
Future work includes demonstrating the effectiveness of \ourMethod on a variety of systems, including but not limited to a changing environment. 
A limiting factor of our approach is the agent does not necessarily take the shortest path in the environment due to the zero-shot nature of our solution. An extension to this work could investigate encoding region proximity into the TS 
or other methods for evaluating the choice of transitions.

\bibliography{main}

\end{document}